\documentclass[11pt]{article}

\PassOptionsToPackage{table}{xcolor}
\PassOptionsToPackage{hyperfootnotes=false}{hyperref}

\usepackage[final]{acl}

\usepackage{times}
\usepackage{latexsym}
\usepackage[T1]{fontenc}
\usepackage[utf8]{inputenc}
\usepackage{microtype}
\usepackage{inconsolata}
\usepackage{graphicx}

\usepackage{booktabs}
\usepackage{amsfonts}
\usepackage{nicefrac}
\usepackage{tabularx}
\usepackage{multirow}
\usepackage{amsmath}
\usepackage{array}
\usepackage{pifont}
\usepackage{makecell}
\usepackage{colortbl}
\usepackage[most]{tcolorbox}
\usepackage{enumitem}
\usepackage{listings}
\usepackage{subfig}
\usepackage{float}
\usepackage{balance}

\usepackage{cleveref}

\crefname{section}{Sec.}{Sec.}
\crefname{figure}{Fig.}{Fig.}
\crefname{table}{Table}{Table}
\crefname{algorithm}{Algorithm}{Algorithm}
\crefname{equation}{Eq.}{Eq.}
\crefname{appendix}{\textbf{Appendix}}{\textbf{Appendix}}
\crefformat{subsection}{Sec.~\thesection-#2#1#3}

\newcommand{\systemname}{little m}

\newlength\savewidth

\definecolor{maroon}{cmyk}{0,0.1,0.01,0.01}
\definecolor{blue}{cmyk}{0.95,0.0,0.2,0.2}
\definecolor{yellow}{cmyk}{0.01,0.0,0.2,0.01}
\definecolor{lightblue}{cmyk}{0.1,0.0,0.02,0.02}
\definecolor{case_verb}{HTML}{fbde84}
\definecolor{case_adj}{HTML}{cccdff}
\definecolor{case_noun}{HTML}{bfeaf1}
\definecolor{case_ff}{HTML}{e65352}
\definecolor{case_error}{HTML}{ffff00}
\definecolor{darkgreen}{RGB}{51,181,41}
\definecolor{darkorange}{RGB}{252,135,62}
\definecolor{t_green}{HTML}{f1f2e4}
\definecolor{LIGHT_BLUE}{HTML}{cce4fe}
\definecolor{LIGHT_RED}{HTML}{f1b9b8}
\definecolor{LIGHT_YELLOW}{HTML}{f1f58a}
\definecolor{LIGHT_GREEN}{HTML}{f1f2e4}
\definecolor{LIGHT_PURPLE}{HTML}{b6a7b9}
\definecolor{lightgray}{gray}{0.95}

\setlist[itemize]{leftmargin=*}
\setlist[enumerate]{leftmargin=*}

\tcbset{
  aibox/.style={
    top=10pt,
    colback=white,
    colframe=black,
    colbacktitle=black,
    enhanced,
    center,
    attach boxed title to top left={yshift=-0.1in,xshift=0.15in},
    boxed title style={boxrule=0pt,colframe=white,},
  }
}
\newtcolorbox{AIbox}[2][]{aibox, title=#2,#1}

\newtcolorbox{promptbox}[2][]{%
  enhanced,breakable,
  colback=gray!5, colframe=black, fontupper=\scriptsize\ttfamily,
  title={\figurename~\refstepcounter{figure}\thefigure: #2},
  #1
}

\title{
\systemname: An AI Agent for Industrial Process Optimization
}

\author{
  Yongchao Ye$^{1}$ \quad
  Xinyu He$^{1}$ \quad
  Dutliff Boshoff$^{1}$ \quad
  Way Kuo$^{1,2}$ \quad
  Lishuai Li$^{1}$\thanks{Corresponding author.} \\
  $^{1}$Department of Data Science, City University of Hong Kong, Hong Kong SAR, China \\
  $^{2}$Hong Kong Institute for Advanced Study, City University of Hong Kong \\
  \texttt{\{yongchao.ye, x.y.he, dboshoff2-c\}@my.cityu.edu.hk} \\
  \texttt{\{way, lishuai.li\}@cityu.edu.hk}
}

\begin{document}
\maketitle

\begin{abstract}

Manufacturing consumes one third of global energy and still has significant room for improvement in terms of energy efficiency. 
Optimal process control is essential for this purpose. 
However, synthesizing mathematical optimization models from messy, real-world industrial specifications requires bridging unstructured natural language and spatial diagrams with rigorous mathematical syntax.
This poses a profound challenge for general-purpose Large Language Models (LLMs), which may introduce invalid constraints when tasked with modeling continuous multi-physics dynamics.
To address this, we introduce \systemname, an AI agent designed to assist the formulation of industrial process control models.
Combining a domain-specific knowledge repository with LLM-driven interaction, the proposed framework formulates real-world optimization problems as mathematical models.
For systematic evaluation, we introduce the Industrial Process Control Benchmark (IPC-Bench), a novel multimodal dataset of 50 canonical scenarios requiring joint reasoning over text and process diagrams.
Through comprehensive automated structural assessments and double-blind human evaluation, \systemname{} substantially outperforms state-of-the-art LLMs, generating semantically correct models.
These evaluations assess formulation quality rather than solver feasibility, formal physical validity, or closed-loop industrial performance.
The implementation of \systemname{} and the IPC-Bench dataset are available at \url{https://github.com/yeyongchao/process-modeling-benchmark}.

\end{abstract}

\section{Introduction}

Large Language Models (LLMs) have demonstrated exceptional capabilities across general natural language tasks, driving interest in their application to complex scientific reasoning and mathematical auto-formalization \cite{zhang2024consistent}. 
Within this scope, industrial process optimization has emerged as a formidable frontier for AI evaluation, given its critical importance to sectors such as energy, manufacturing, and pharmaceuticals \cite{ olsson2023near}. 
Unlike basic arithmetic word problems or general code generation, solving real-world industrial problems requires translating unstructured text and spatial topologies into rigorous mathematical models governed by multi-physics phenomena and non-linear dynamics \cite{xu2024transparent}. 
Modeling such continuous systems demands a level of structural prediction and semantic grounding that challenges the inherent limits of purely autoregressive generation \cite{wu2025training}.
For example, a useful formulation must distinguish decision variables from measurements and fixed parameters because these roles determine model interpretation.

Consequently, research has actively transitioned toward specialized LLM agents for Operations Research (OR), utilizing techniques like structured fine-tuning, multi-agent decomposition, and solver-in-the-loop verification to synthesize complex logic \cite{jiang2024llmopt, tran2025multi, zhang2025or}.
However, while these advanced frameworks show promise in general optimization tasks like logistics and linear programming, they expose critical vulnerabilities when applied to continuous multi-physics processes. 
Without explicit process grounding, monolithic LLMs may generate mathematically plausible but physically inconsistent constraints.
Directly automating industrial modeling with these tools is difficult because the task is multimodal and layered with tacit physical heuristics, such as thermodynamic feasibility, that are rarely detailed in problem descriptions \cite{Le2025making, qu2025hierarchical}.

Furthermore, existing AI-driven industrial solutions, including digital twins and control-theoretic foundation models, predominantly focus on simulation or parameter tuning within fixed infrastructure rather than the de novo synthesis of mathematical models from ambiguous intents \cite{tao2024advancements, maher2025llmpc}. 
This gap is reflected in current benchmarks. Standard multimodal AI datasets often evaluate mathematical reasoning as purely abstract logic or static visual geometry \cite{huang2025orlm, xiao2023chain}. 
However, real-world industrial formulation requires explicit treatment of physical and operational constraints followed by engineering validation.
This disparity motivates auditable AI architectures in which neural language models are coupled with structured domain knowledge and expose intermediate decisions for human review \cite{raspanti2025grammar}.

\begin{table*}
    \centering
    \small
    \caption{Comparison of mathematical and optimization modeling benchmarks.}
    \label{tab:benchmark_comparison}
    \resizebox{\textwidth}{!}{
    \begin{tabular}{lp{3.5cm}p{3cm}p{3cm}p{4cm}c}
        \toprule
        \textbf{Benchmark} & \textbf{Domain / Problem Type} & \textbf{Input Modality} & \textbf{Output} & \textbf{Evaluation} & \textbf{Problems} \\
        \midrule
        ComplexOR \cite{xiao2023chain} & OR / MILP, LP & Text & Solver code (Gurobi/Python) & Execution-based: Objective vs ground truth & 37 \\
        NL4OPT \cite{ramamonjison2023nl4opt} & LP & Text & Designed template & Extraction/Generation: F1 \& syntax & 1101 \\
        IndustryOR \cite{huang2025orlm} & Logistic \& scheduling / LP, MIP, NLP & Text + Tabular data & Math model, code & Execution-based: Involve human experts & 100 \\
        MAMO \cite{huang2025llms} & ODEs, LP & Text & Solver code (Python) & Execution-based: Accuracy & 211 \\
        GSM8K \cite{cobbe2021training} & Arithmetic Word Problems & Text & Numerical Value & Accuracy: exact match & 8500 \\
        \rowcolor{gray!10} \textbf{IPC-Bench (Ours)} & Process Control & \textbf{Text + Process Diagrams} & \textbf{Math Model} & \textbf{Semantic \& Structural Grounding} & \textbf{50} \\
        \bottomrule
    \end{tabular}
    }
\end{table*}

To overcome this fundamental barrier, we introduce \systemname, an AI agent designed to assist industrial process optimization model formulation.
\systemname{} moves beyond the limitations of general-purpose AI by integrating two distinct components:
(1) A structured knowledge repository covering process-optimization directions, control strategies, modeling templates, and recurring constraint patterns; and
(2) An LLM that serves as an interactive interface, structures user-provided information, retrieves relevant entries, and exposes intermediate outputs for review.
Together, these components produce structured optimization models grounded in supplied process information and retrieved domain knowledge.
In this work, we evaluate this grounded, multimodal architecture on IPC-Bench, which focuses on process-industry cases governed by physical mechanisms shared across continuous-process sectors.
Our primary contributions are:
\begin{itemize}

\item The architecture and implementation of \systemname, a knowledge-grounded assistant for industrial process control model formulation, combining domain-specific knowledge with LLM-driven interaction.

\item The creation of Industrial Process Control Benchmark (IPC-Bench), an initial benchmark comprising 50 canonical, textbook-derived process-industry optimization scenarios.
Unlike existing unimodal datasets, IPC-Bench requires reasoning over multimodal inputs to capture spatial and topological contexts.

\item A comprehensive validation combining automated structural assessments with a double-blind human evaluation.
Results show higher expert preference and better formulation quality than the evaluated Qwen3 and DeepSeek baselines.

\end{itemize}

\section{Related Work}

\subsection{LLM Agents for Optimization and Operations Research}
The application of Large Language Models (LLMs) to Operations Research (OR) has transitioned from elementary prompt engineering to sophisticated, agentic frameworks. 
To address the strict syntactical and logical demands of exact solvers, recent literature emphasizes structured fine-tuning (e.g., LLMOPT \cite{jiang2024llmopt}), modular decomposition (e.g., OptiMUS \cite{ahmaditeshnizi2023optimus}), and multi-agent collaboration (e.g., Chain-of-Experts \cite{xiao2023chain}). 
A notable paradigm shift is the adoption of closed-loop reasoning, which embeds optimization solvers into the training or deployment loop as objective verifiers to reinforce mathematically valid reasoning paths \cite{zhang2025or, chen2025solver}. 
Despite these advances in general OR, industrial process control poses unique challenges due to its inherent reliance on non-linear continuous dynamics and differential equations \cite{fakih2024llm4plc}. 
While emerging frameworks like ControlAgent \cite{guo2024controlagent} and LLMPC \cite{maher2025llmpc} have integrated LLMs for control parameter tuning and high-level trajectory planning, a unified architecture capable of interactively eliciting physically grounded optimization formulations remains underexplored.

\subsection{Benchmarking and Evaluation Paradigms}
The development of optimization agents has necessitated specialized evaluation frameworks, a selection of which is compared in \cref{tab:benchmark_comparison}. 
Initial benchmarks, such as NL4OPT \cite{ramamonjison2023nl4opt}, established baselines for linear programming extraction. 
Subsequent datasets like IndustryOR \cite{huang2025orlm} and ComplexOR \cite{xiao2023chain} introduced large-scale logistics and scheduling scenarios. 
However, a significant benchmarking gap persists regarding industrial process control. 
Although datasets like MAMO \cite{huang2025llms} include ordinary differential equations, they largely treat them as abstract mathematics rather than intersecting continuous physics with discrete operational constraints. 
Consequently, execution rates and semantic similarity capture useful aspects of formulation quality, but neither alone establishes physical validity or operational safety for deployment.


\subsection{AI-Assisted Process Optimization}

Process systems engineering connects process modeling and design with operations optimization and control \cite{grossmann2019pse}. 
At the operational level, RTO computes economic targets using updated steady-state models, whereas MPC performs finite-horizon dynamic optimization for constrained control \cite{darby2011rto}. 
Industrial AI complements these model-based methods through digital twins, physics-informed learning, and process-industry foundation-model architectures \cite{tao2024advancements,liu2025physics,ren2025foundation}.
Recent AI-assisted engineering systems address control-structure prediction from process diagrams, externally verified PLC code generation, and computation-in-the-loop controller design, bringing language models closer to concrete engineering artifacts \cite{fakih2024llm4plc,guo2024controlagent}. 
Other frameworks use structured prompting, engineering tools, or process simulators to support flowsheet analysis, simulation, and optimization through iterative task decomposition \cite{tao2025pse,zeng2025llm}. 
Together, these studies reflect a broader shift toward structured and tool-supported engineering assistance.

\section{Preliminary}

We define the task of automated industrial modeling as a conditional generation problem mapping a multimodal context $\mathcal{X}$ to a formal optimization model $\mathcal{M}$.
The input space $\mathcal{X} = \{D_{\textnormal{text}}, D_{\textnormal{img}}\}$ encapsulates the unstructured engineering intent, where $D_{\textnormal{text}}$ comprises natural language narratives detailing operational goals and $D_{\textnormal{img}}$ denotes visual process topologies.
The target output is a structured triplet $\mathcal{M} = (\mathcal{V}, \mathcal{F}, \mathcal{C})$, representing a canonical optimization model.
Here, $\mathcal{V}$ defines the set of decision variables grounded in physical domains (e.g., mass flows $f \in \mathbb{R}_{\ge 0}$, binary actuation states $z \in \{0,1\}$); $\mathcal{F}$ represents the scalar or vector-valued objective function (e.g., minimization of energy cost); and $\mathcal{C}$ constitutes the set of equality and inequality constraints representing mass/energy balances, thermodynamic limits, and safety interlocks.

This task differs from standard code generation because process connectivity and physical relations must be represented explicitly.
For example, a candidate constraint set should respect the connectivity graph in $D_{\textnormal{img}}$ and include applicable conservation relations such as $\sum m_{in}=\sum m_{out}$ at a steady-state node.
We treat these as formulation requirements to be audited, not as properties formally guaranteed by the generator.
The agent approximates the expert mapping $f^*: \mathcal{X} \rightarrow \mathcal{M}$ and returns a candidate specification for review and subsequent numerical implementation.
Our evaluation compares $\mathcal{V}$, $\mathcal{F}$, and $\mathcal{C}$ with expert-written references using structural metrics and expert judgment.
It does not execute the model in a solver or validate closed-loop operation.

\section{Methodology}

\subsection{System Architecture}\label{sec:design}

\begin{figure*}
    \centering
    \includegraphics[width=0.9\linewidth]{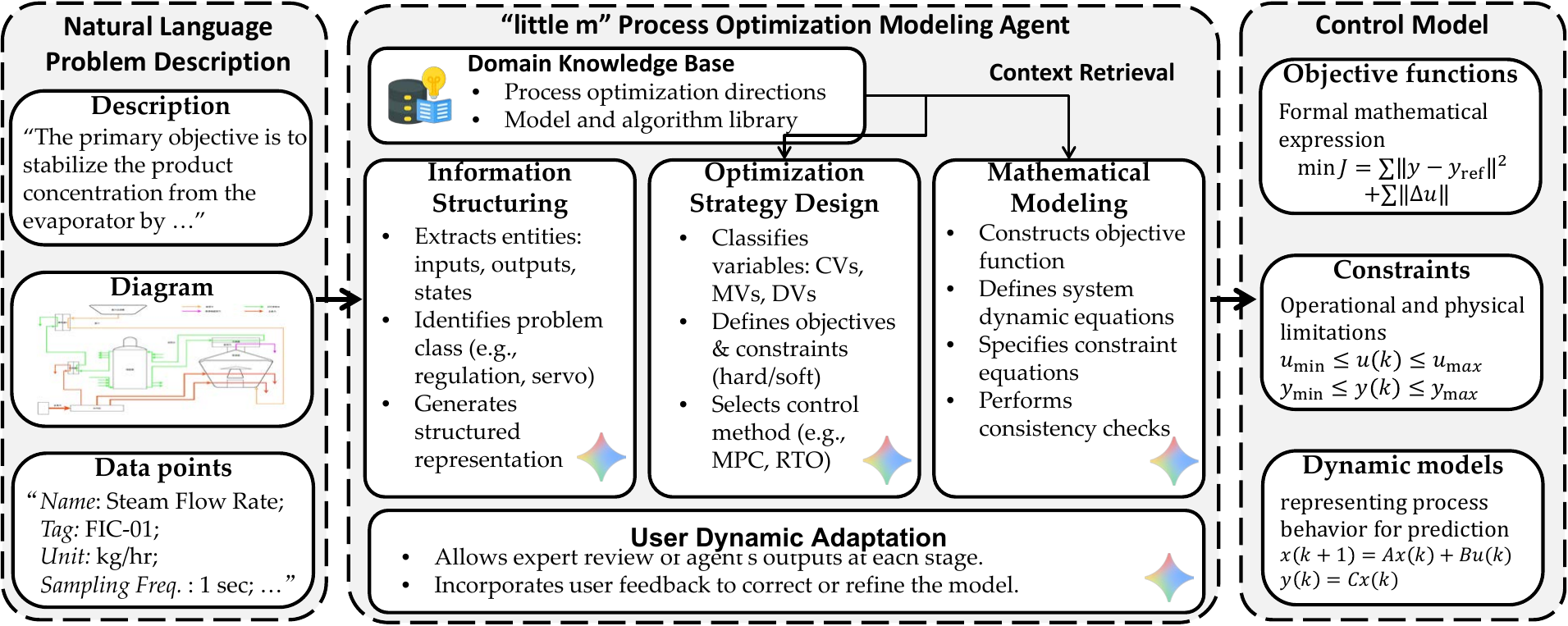}
    \caption{Architecture of \systemname. The system employs a three-stage cognitive pipeline driven by an LLM-RAG core to ground optimization synthesis in domain-specific knowledge.}

    \label{fig:workflow}
\end{figure*}

The design of \systemname{} is illustrated in \cref{fig:workflow}. 
Instead of attempting a direct, single-step translation from $\mathcal{X}$ to $\mathcal{M}$, we implement a hierarchical cognitive workflow that mirrors the standard project life cycle of human process engineers. 
The motivation for this multi-stage design is twofold. 
First, industrial problem statements are inherently ambiguous and often rely on ``tacit knowledge'' that is not explicitly stated in the prompt, e.g., assuming a tank cannot overflow.  
A direct translation model frequently hallucinates invalid constraints or misses these implicit safety bounds.
Second, real-world engineering projects invariably follow a sequential structure: engineers first validate their understanding of the process flow, then determine the control strategy, and only then derive the specific equations.  
This structure not only improves model accuracy but also facilitates the user dynamic adaptation loop, allowing domain experts to intervene and refine the strategy before the complex mathematical syntax is generated.

The reasoning engine operates through a strictly sequential three-stage pipeline. 
The initial stage, \textit{Information Structuring}, functions as an intake filter to handle the ambiguity of raw engineering intent. Then, in \textit{Strategy Design}, the agent acts as a lead engineer identifying optimization opportunities.  The final stage of the reasoning core, \textit{Mathematical Modeling}, translates this confirmed strategy into a formal formulation $\mathcal{M}$.

\paragraph{Information Structuring}

The initial stage, Information Structuring, involves gathering and organizing all relevant information provided by the user to handle the ambiguity of raw engineering intent. 
We formulate this stage as a state extraction function $S_{0} = \Phi_{extract}(\mathcal{X})$ that compresses the raw multimodal input $\mathcal{X}$ into a structured representation. 
The primary input for this stage includes textual narratives of the production process and visual functional diagrams. 
Relying on this user-provided information, \systemname{} structures the raw data into a coherent summary $S_{0}$, capturing key steps, equipment types, and operational modes. 
A human-in-the-loop checkpoint concludes this stage.
Defining a user acknowledgement function $V_{user}(s_t) \in \{0, 1\}$, the transition to the next state requires $V_{user}(S_{0}) = 1$.
This exposes omissions for correction but does not certify that $S_0$ is complete or correct.

\paragraph{Strategy Design}

Based on the structured state $S_{0}$, the user's objective, and retrieved knowledge $K_{ret}$, Strategy Design generates a high-level control strategy $Z=G_{\theta}(S_0,K_{ret})$.
Here, \systemname{} identifies optimization opportunities and formulates a high-level control strategy.
The workflow invokes an LLM to generate the strategy and passes the structured output to the mathematical modeling stage.
The strategy specifies the optimization goal, model class, primary decision variables, information requirements, and candidate constraint families.
The transition to mathematical formulation occurs after the user acknowledges $Z$.
This gate provides an opportunity for correction rather than a mathematical restriction on hallucination.

\paragraph{Mathematical Modeling}

In the final stage, Mathematical Modeling, \systemname{} translates the acknowledged strategy into a formal mathematical model.
The generation step is $\mathcal{M}=H_{\theta}(S_{0},Z,K_{ret})$, showing that the formulation is conditioned on the structured state, strategy, and retrieved knowledge.
The output separates objective functions $\mathcal{F}$, decision variables $\mathcal{V}$, fixed parameters, and constraints $\mathcal{C}$, with a data-source field for each symbol.
The final checkpoints are structured review artifacts.
They ask whether each diagram connection is represented, units and symbol definitions are consistent, required operating bounds are present, and the proposed model class is compatible with a named solver family.
They improve auditability but do not execute the candidate model, prove topological integrity, or establish numerical feasibility.

\paragraph{Iterative Refinement and Adaptation}
The final stage introduces the critical feedback loop that distinguishes \systemname{} from static translation tools. 
Recognizing that the initial model may still misalign with the user's unstated intent or tacit knowledge, this stage formalizes the Interactive Elicitation process. 
The system presents the generated model to the domain expert for semantic review.
If discrepancies are identified—for example, if a constraint violates a specific start-up procedure, the agent triggers a multi-turn dialogue. 
This dialogue is not merely a chat but a targeted investigation where the agent proposes hypotheses and the user provides corrections. 
These corrections are fed back into the system, propagates changes through the formulation layers. 
This cycle continues until the user accepts the candidate for subsequent engineering and numerical validation.

\begin{figure*}
    \centering
    \includegraphics[width=0.9\linewidth]{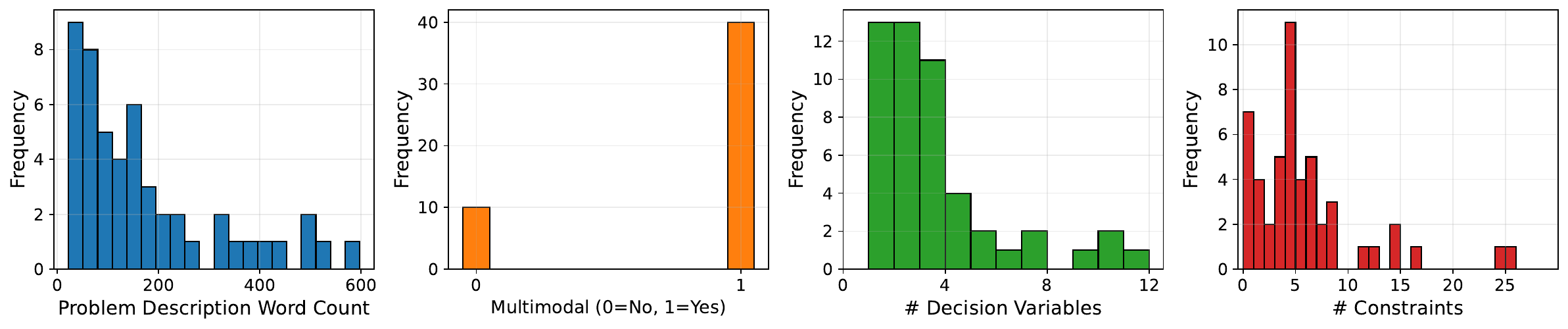}
    \caption{Metadata distribution of the IPC-Bench dataset.}
    \label{fig:benchmark_metadata}
\end{figure*}

\paragraph{Domain Knowledge Grounding}
To support this reasoning process, the agent is grounded by a domain-specific retrieval system that bridges the gap between general linguistic competence and specific industrial expertise.
We organize the curated knowledge base as self-contained problem-path entries rather than isolated algorithm descriptions. 
Each entry connects an observable industrial symptom or operating context to a primary optimization objective, conditional candidate methods, required process information, reusable mathematical and constraint patterns, and applicability boundaries.
The repository combines general pathways that recur across process industries with contextualized entries distilled from our industrial project experience. 
A symptom is treated as a retrieval anchor rather than a confirmed root cause, and a candidate method is proposed together with the conditions and information needed to justify its use.
Technically, a query-augmentation module maps the current process context and objective to relevant entries through vector search, and a cross-encoder reranker selects the context $K_{ret}$ supplied to the LLM. 
The retrieved entries support Strategy Design with candidate targets, methods, and missing-information cues, and support Mathematical Modeling with reusable formulation patterns and explicit boundaries. 
They provide contextual engineering guidance rather than numerical plant models, automatically enforceable physical laws, or solver-level validation; further details are provided in \cref{appendix:kb}.

\subsection{Implementation}

We implemented a modular prompting strategy governed by finite-state workflow logic with role-constrained instructions for each stage: Information Structuring employs gap analysis, Strategy Design formulates a high-level control strategy with user confirmation, and Mathematical Formulation imposes data-variable mapping rules that request a source for every symbol.
The system uses Gemini 2.5 Pro as the reasoning engine, with BAAI's bge-m3 for embeddings and bge-reranker-v2-m3 for retrieval precision.
Detailed prompt templates are provided in \cref{appendix:prompts}, and an end-to-end trace of the intermediate representations is shown in \cref{appendix:cases}.

\section{IPC-Bench Dataset}

To evaluate formulation on representative process-industry problems, we constructed IPC-Bench from textbooks in process control and chemical engineering optimization \cite{textbook2001optimization, textbook2016process, leblanc2009process, kookos2022practical, biegler2010nonlinear, rao2010advanced, agachi2016advanced, ingham2008chemical}.
We selected scenarios involving features such as nonlinear constraints, dynamic relations, process diagrams, and mixed-integer variables.
As highlighted in \cref{tab:benchmark_comparison}, IPC-Bench evaluates semantic and structural agreement of mathematical formulations synthesized from text and process diagrams.
Although IPC-Bench is an initial benchmark of 50 cases, its core unit operations involve thermodynamics, fluid mechanics, reaction kinetics, and mass and heat transfer, which are physical mechanisms shared across continuous-process sectors such as metallurgy, petrochemicals, pharmaceuticals, and food processing.

\paragraph{Dataset Characteristics}
The dataset consists of 50 cases covering reaction, separation, thermal, and utility/scheduling scenarios, from utility cost tracking to reactor yield maximization.
The dataset presents varying complexity: most problems require multimodal inputs (text + process flow diagrams), with the most challenging scenarios demanding numerous decision variables and extensive, highly coupled constraint sets.
Each problem is decomposed into two components: 
(1) Problem Description preserving raw context (narrative, operational logic, system schematics), 
and (2) Ground Truth Model with explicit segmentation into decision variables, objective functions, and constraints, each annotated with mathematical symbols, physical domains, and natural language descriptions.
This structure enables fine-grained evaluation of whether an agent correctly identifies physical entities and formulates governing equations. 
Detailed data structure examples are provided in \cref{appendix:dataset}.
Because the cases are textbook-derived, potential exposure during model pretraining cannot be excluded. Building a complementary real-world industrial benchmark remains future work and requires data standardization, proprietary-information redaction, and evaluation criteria established through domain-expert consensus.


\begin{table*}[t]
    \centering
    \caption{Comparative Analysis (Binomial Test, $H_0: p=1/3$). Win Rates indicate the percentage of expert votes. Significance levels ($^* p<0.05$, $^{**} p<0.01$, $^{***} p<0.001$). }
    \label{tab:model_comparison}
    \begin{tabular}{l c c c c c c}
        \toprule
        \multirow{2}{*}{\textbf{Evaluation Aspect}} & \multicolumn{2}{c}{\textbf{Qwen3}\textsuperscript{\dag}} & \multicolumn{2}{c}{\textbf{DeepSeek}\textsuperscript{\ddag}} & \multicolumn{2}{c}{\textbf{\systemname}} \\
        \cmidrule(lr){2-3} \cmidrule(lr){4-5} \cmidrule(lr){6-7}
        & \textbf{win rate} & \textbf{$p$-value} & \textbf{win rate} & \textbf{$p$-value} & \textbf{win rate} & \textbf{$p$-value} \\
        \cmidrule(lr){1-7}

        Objective Function    & 16.0\%  & $0.010^{**}$  & 26.0\%  & $0.30$  & \textbf{58.0\%}  & $\mathbf{<0.001^{***}}$ \\
        Decision Variables    & 14.0\%  & $0.003^{**}$  & 26.0\%  & $0.30$  & \textbf{60.0\%}  & $\mathbf{<0.001^{***}}$ \\
        Constraints           & 22.0\%  & $0.099$  & 26.0\%  & $0.30$  & \textbf{52.0\%}  & $\mathbf{0.007^{**}}$ \\
        Overall Quality       & 16.0\%  & $0.010^{**}$  & 18.0\%  & $0.024^{*}$  & \textbf{66.0\%}  & $\mathbf{<0.001^{***}}$ \\

        \bottomrule
        \multicolumn{7}{l}{\footnotesize \textsuperscript{\dag} Qwen3: Qwen3-Next-80B-A3B-Instruct; \textsuperscript{\ddag} DeepSeek: Deepseek-V3.2}

    \end{tabular}
\end{table*}

\section{Experiments}

\subsection{Experimental Setup}

We evaluated \systemname{} against Qwen3-Next-80B-A3B-Instruct and DeepSeek-V3.2.
We designed a dual-perspective evaluation framework because standard approaches are fundamentally insufficient for industrial optimization. NLP metrics fail to recognize algebraic equivalencies without lexical overlap, and execution-based metrics are impractical when ready-made digital twins are unavailable.
Our framework combines: (1) double-blind human-expert assessment of practical utility and physical logic across 20 benchmark cases, and (2) automated machine-based evaluation measuring structural accuracy against expert-verified ground truth across all 50 cases.

\subsection{Human Evaluation Design}

The human evaluation was conducted via a structured survey instrument.
The evaluation panel consisted of 8 domain specialists (Ph.D. students and postdocs) from complementary disciplines: 3 from computer science, 2 from data science, and 3 from control engineering.
Each problem was evaluated by 2--4 independent experts.

Experts assessed each blinded, randomized candidate across four dimensions:
(1) Objective Function Quality---algebraic correctness and fidelity to the stated target;
(2) Decision Variable Completeness---coverage of physical quantities with appropriate domains;
(3) Constraint Validity---absence of unsupported or topologically inconsistent conditions;
and (4) Overall Convincingness---holistic usefulness as a candidate formulation for further engineering.
All models were strictly anonymized and presentation order randomized to mitigate bias.
Detailed survey methodology is provided in \cref{appendix:survey}.

\subsection{Machine Evaluation Design}

To complement human assessment with scalable, reproducible metrics, we developed an automated evaluation pipeline that compares the predicted model $M_p = \{V_p, f_p, C_p\}$ against the ground truth $M_g = \{V_g, f_g, C_g\}$.

\paragraph{Decision Variables.}
Variables are evaluated as sets. A regex module normalizes variable names, and an LLM-driven semantic mapping aligns differently named but physically equivalent entities. Structural fidelity is measured via Jaccard similarity:
\begin{equation}
    S_{vars} = \frac{|V_g \cap V_p'|}{|V_g \cup V_p'|}
\end{equation}

\paragraph{Objective Function.}
The evaluation script strips natural language noise (prefixes like ``Minimize'', ``Max'') and computes token-level Jaccard similarity:
\begin{equation}
    S_{obj} = \frac{|T_g \cap T_p|}{|T_g \cup T_p|}
\end{equation}

\paragraph{Constraint Set.}
All terms are normalized to standard form ($A \le B \to A - B \le 0$). A bipartite matching algorithm (Hungarian) pairs each generated constraint with its closest ground truth equivalent based on token similarity weights $w_{ij}$. Scoring uses a continuous F1 metric:
\begin{equation}
    P = \frac{\sum w_{ij}}{|C_p|}, \quad R = \frac{\sum w_{ij}}{|C_g|}, \quad S_{cons} = 2 \cdot \frac{P \cdot R}{P + R}
\end{equation}
Precision penalizes hallucinated constraints; recall penalizes missing safety interlocks.
Full protocol details are provided in \cref{appendix:machine-eval}.

\subsection{Main Results}

\subsubsection{Human Evaluation}

\cref{tab:model_comparison} presents expert preference win rates tested against a random baseline ($H_0: p=1/3, N= \textnormal{received evaluations}$).
Among the three candidates included in the survey, \systemname{} received the highest preference rate across all categories: Objective Function (58.0\%), Decision Variables (60.0\%), Constraints (52.0\%), and Overall Quality (66.0\%), with statistically significant deviations from chance ($p < 0.01$ for all dimensions).
The 60.0\% preference rate for decision variables ($p < 0.001$) is consistent with the system's use of structured variable tables aligned with data points, although the survey does not isolate that component causally.
In the survey, Qwen and DeepSeek received decision-variable preference rates of 14.0\% ($p = 0.003^{**}$) and 26.0\% ($p = 0.30$), respectively, compared with 60.0\% for \systemname{}.

\begin{figure}
    \centering
    \includegraphics[width=\linewidth]{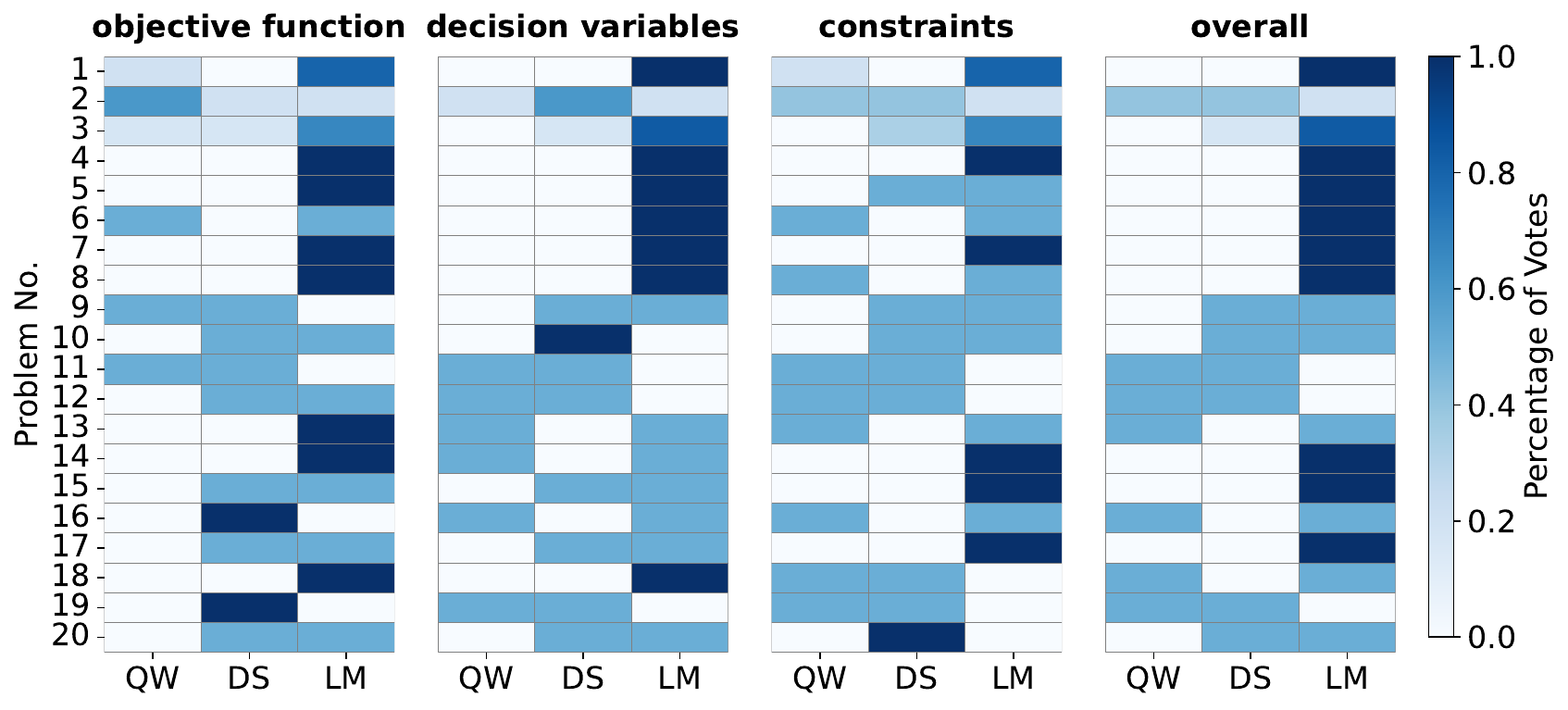}

    \caption{Heatmap of expert consensus (win rate) across modeling dimensions for Qwen3 (QW), Deepseek (DS), and \systemname{} (LM). Darker cells indicate higher agreement.}

\label{fig:heatmap}
\end{figure}

\cref{fig:heatmap} visualizes expert consensus across modeling dimensions.
The Decision Variables column displays high-consensus patterns, frequently achieving expert unanimity in the evaluated cases.
Consensus for Objective Functions and Constraints was more heterogeneous, reflecting inherent ambiguity in translating abstract control goals into mathematical terms, yet \systemname{} maintained the highest aggregate preference.

Analysis of the responses indicates that domain experts diverge on stylistic conventions such as constraint grouping, while more often agreeing on clear omissions or unsupported relations.
This motivates evaluation protocols that accommodate equivalent formulations while still penalizing structural errors.

\subsubsection{Machine Evaluation}

\cref{tab:machine_evaluation} reports machine-based structural scores.
Among the three evaluated systems, \systemname{} obtains the highest Decision Variable score (0.733) and Constraint score (0.418).
DeepSeek obtains the highest objective score (0.553).
These reference-alignment metrics therefore do not support a broad claim that the staged system improves every final-formulation dimension; the subsequent interaction and ablation studies examine narrower effects.

\begin{table}
    \centering
    \small
    \caption{Machine-based evaluation scores.}
    \begin{tabular}{lccc}
    \toprule
    \textbf{Aspects} & \textbf{Qwen3} & \textbf{Deepseek} & \textbf{\systemname} \\
    \midrule
    Decision Variables & 0.673 & 0.699 & \textbf{0.733} \\
    Objective Functions & 0.526 & \textbf{0.553} & 0.518 \\
    Constraints & 0.389 & 0.395 & \textbf{0.418} \\
    \bottomrule
    \label{tab:machine_evaluation}
\end{tabular}
\end{table}

\begin{figure*}
    \centering
    \includegraphics[width=\linewidth]{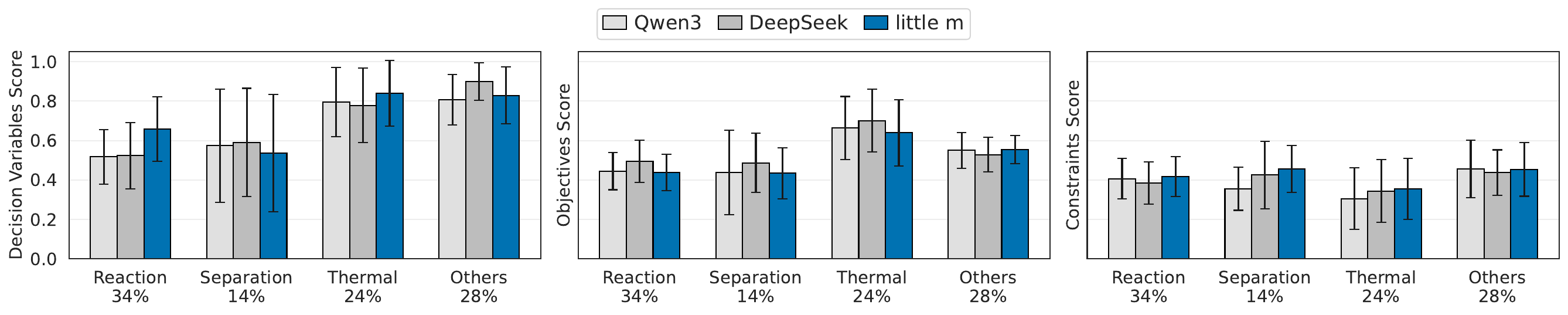}
    \caption{Performance across industrial process categories, displaying machine-evaluated scores for Decision Variables, Objective, and Constraints.}
    \label{fig:sym_eval_group_category}
\end{figure*}

\begin{figure}
    \centering
    \includegraphics[width=\linewidth]{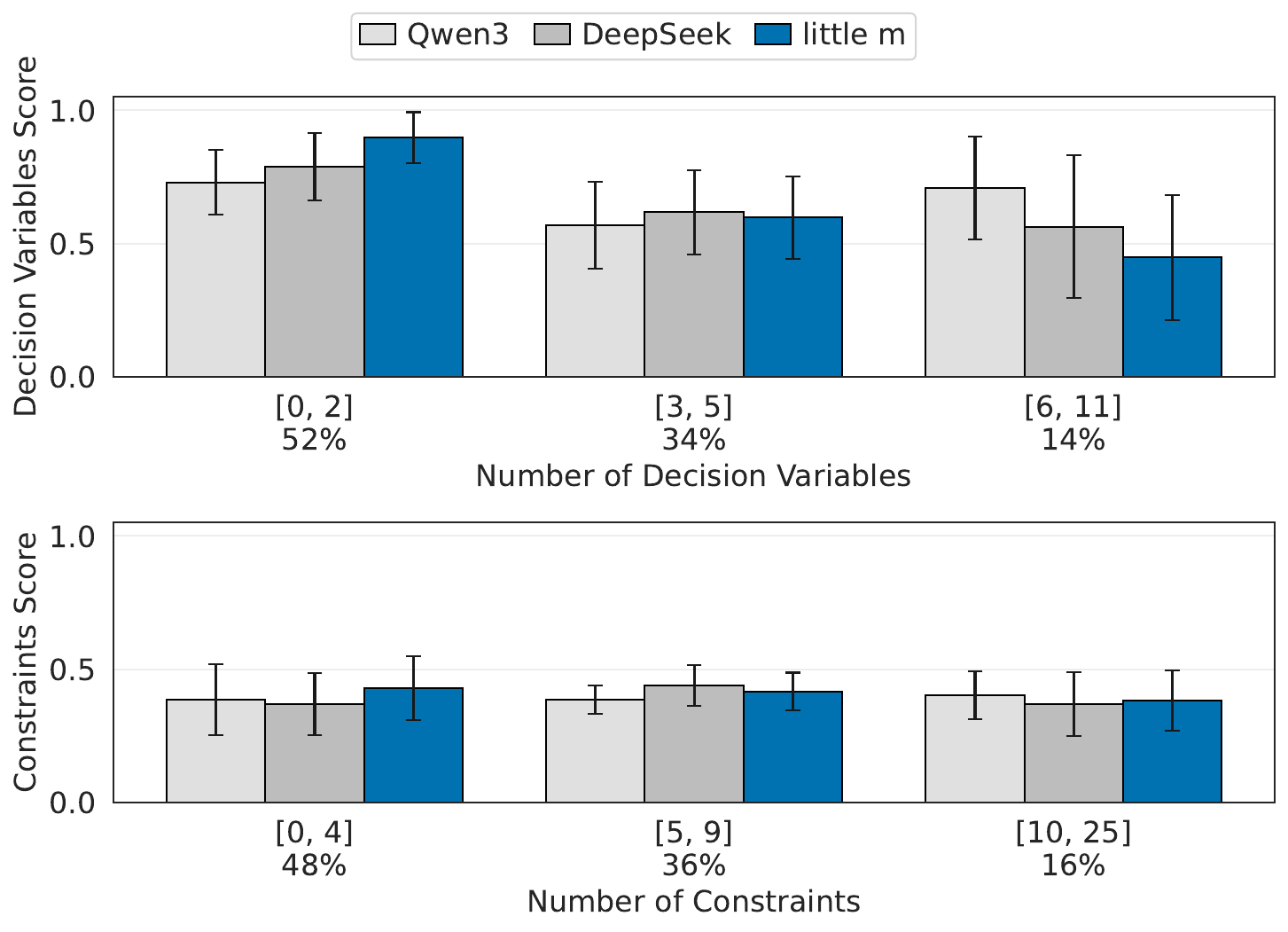}
    \caption{Machine evaluation scores stratified by condition complexity: decision variable accuracy (top) and constraint formulation accuracy (bottom). Error bars indicate variance within each complexity bin.}
    \label{fig:sym_eval_group_case_statis}
\end{figure}

\subsubsection{Human-Machine Alignment}

To estimate whether the automated metrics track expert preference within the original three-system comparison, we cross-validated machine scores against human evaluations.
For each of the 20 evaluated cases, the machine pipeline ranked candidate models and we quantified alignment using Top-1 Accuracy (frequency the machine's top choice matches human preference) and Mean Reciprocal Rank: $\text{MRR} = \frac{1}{N} \sum_{i=1}^{N} \frac{1}{\text{rank}_i}$, where $\text{rank}_i$ is the position of the human-preferred model in the machine-sorted list.
As shown in \cref{tab:human_machine_alignment}, Top-1 Accuracy is 0.75 for both Decision Variables and Objective Functions, with MRR scores of 0.867 and 0.858.
Constraint alignment is slightly lower (Top-1: 0.60, MRR: 0.792), reflecting inherent subjectivity in translating control goals into logical constraints.
An end-to-end case trace illustrating how the staged workflow constructs a control model is provided in \cref{appendix:cases}.

\begin{table}
    \centering
    \small
    \caption{Alignment between Machine-based Evaluation and Expert Preference.}
    \label{tab:human_machine_alignment}
    \begin{tabular}{lcc}
        \toprule
        \textbf{Evaluation Aspect} & \textbf{Top-1 Acc} & \textbf{MRR} \\
        \midrule
        Objective Function & 0.75 & 0.858 \\
        Decision Variables & 0.75 & 0.867 \\
        Constraints & 0.60 & 0.792 \\
        \bottomrule
    \end{tabular}
\end{table}

\subsubsection{Results by Complexity and Domain}

To investigate how performance scales with problem complexity, we stratified benchmark problems by the number of decision variables and constraints (\cref{fig:sym_eval_group_case_statis}).
\systemname{} demonstrates a distinct advantage in low-to-medium complexity scenarios, achieving the highest scores for problems with fewer than six variables.
Across all models, constraint formulation remains a persistent bottleneck, with scores persistently lower than variable identification regardless of problem scale.
Nevertheless, \systemname{} exhibits notable stability in constraint formulation even as constraint count increases up to 25.

We further disaggregated evaluation metrics across distinct physical operations to observe domain-specific proficiencies (\cref{fig:sym_eval_group_category}).
The results indicate that Thermal scenarios are generally the easiest to model, yielding the highest average variable and objective scores across all frameworks.
Within the original three-system comparison, \systemname{} scores highest in Reaction and Thermal problems, a pattern consistent with the coverage of thermodynamic and kinetic entries in the knowledge base.
Conversely, DeepSeek demonstrates specialized strength in Separation tasks, while Qwen exhibits the highest variance across engineering categories.

\subsection{Interaction Analysis}

To test how \systemname{} can recover from incomplete requests, we created an incomplete-input condition by withholding information about process structure, operating limits, or control information.
All settings share the same reasoning engine, workflow, machine evaluator, and references:

\begin{itemize}[leftmargin=*,topsep=2pt,itemsep=1pt]
    \item \textbf{Full Input:} The standard \systemname{} setting used in the main results receives the original request and generates the formulation in one shot.
    \item \textbf{Incomplete Input:} \systemname{} receives the masked request and generates the formulation in one shot without clarification.
    \item \textbf{Interactive Recovery:} Starting from the same masked request, \systemname{} asks clarification questions, and a controlled simulator discloses a withheld fact only when a question is relevant to it. This controls disclosure for repeatability but does not reproduce all behavior of an industrial user.
\end{itemize}

As shown in \cref{tab:interaction_recovery}, interactive recovery clearly improves on incomplete input, with especially large gains in decision variables and constraints, confirming the contribution of masked information.
Treating full input as an approximate upper bound, interactive recovery achieves comparable overall performance, scoring higher on decision variables, similarly on constraints, and lower on objective functions.
These results suggest that interaction recovers most of the withheld formulation information, while objective-related details remain harder to recover through clarification.

\begin{table}[H]
    \centering
    \small
    \setlength{\tabcolsep}{2pt}
    \caption{Recovery from incomplete requests under machine-based evaluation.}
    \label{tab:interaction_recovery}
    \begin{tabular}{lccc}
        \toprule
        \textbf{Evaluation Aspect} & \makecell{\textbf{Full} \\ \textbf{Input}} & \makecell{\textbf{Incomplete} \\ \textbf{Input}} & \makecell{\textbf{Interactive} \\ \textbf{Recovery}} \\
        \midrule
        Decision Variables & 0.733 & 0.721 & \textbf{0.802} \\
        Objective Functions & \textbf{0.518} & 0.431 & 0.443 \\
        Constraints & 0.418 & 0.314 & \textbf{0.423} \\
        \bottomrule
    \end{tabular}
\end{table}

\subsection{Ablation Study}

We conduct two single-component ablations of the full \systemname{} pipeline.
The w/o Knowledge variant removes retrieved entries but retains the request and diagram; w/o Diagrams removes the visual input but retains the request and retrieved knowledge.
The variants are evaluated on the same settings.

As shown in \cref{tab:component_ablation}, both ablations reduce performance, confirming that knowledge and diagrams provide complementary information.
Removing knowledge most strongly affects constraint alignment, suggesting that retrieved entries supply domain-specific patterns for formulating operational constraints.
The objective and constraint drops without diagrams indicate that process topology connects identified quantities to goals and dependencies.
Thus, knowledge provides formulation guidance, while diagrams ground it in process topology.

\begin{table}
    \centering
    \small
    \setlength{\tabcolsep}{2pt}
    \caption{Ablations under machine-based evaluation.}
    \label{tab:component_ablation}
    \begin{tabular}{lccc}
        \toprule
        \textbf{Evaluation Aspect} & \textbf{\systemname} & \makecell{\textbf{w/o} \\ \textbf{Knowledge}} & \makecell{\textbf{w/o} \\ \textbf{Diagrams}} \\
        \midrule
        Decision Variables & 0.733 & 0.720 & 0.723 \\
        Objective Functions & 0.518 & 0.510 & 0.468 \\
        Constraints & 0.418 & 0.381 & 0.383 \\
        \bottomrule
    \end{tabular}
\end{table}


\section{Conclusion}

We introduced \systemname, an AI agent for industrial process optimization that bridges the semantic gap between physical descriptions and mathematical formulations through a three-stage cognitive pipeline grounded by domain-specific knowledge retrieval.
On IPC-Bench, \systemname{} received higher expert preference than Qwen3 and DeepSeek and obtained the highest variable and constraint scores among the three systems under machine-based structural metrics.
These results concern candidate formulation quality and do not establish solver feasibility, formal physical validity, or closed-loop performance.
Future work will integrate \systemname{} with numerical solvers and evaluate broader industrial cases and task-specific workflows.

\section{Limitations}

Several limitations remain that suggest directions for future work.
First, extending the curated knowledge base requires expert curation for each target domain, limiting domain-agnostic scalability.
Second, the system lacks solver-level verification, so numerical issues may remain undetected until implementation.
Third, IPC-Bench contains only 50 textbook-derived cases.
Broader validation requires independently collected industrial cases, and potential pretraining exposure cannot be excluded.
Finally, the machine metrics may penalize equivalent formulations, and the ablations do not isolate every workflow component.

\section{Ethical Considerations}

\systemname{} is designed as a formulation assistant, not an autonomous decision-maker; all generated models require human review before deployment in safety-critical industrial processes.
The retrieval corpus may reflect biases present in its source literature, which could skew formulations toward well-documented process types and underserve novel or under-represented domains.

\section*{Acknowledgments}

This work was supported by City University of Hong Kong under Grant PJ9361031 and Baosteel--City University of Hong Kong Joint Research Centre under Grant BHK2502-01.

We thank the domain experts who participated in our evaluation.
LLMs assisted with language polishing and preliminary cleaning of textbook materials used to construct IPC-Bench.

little m is dedicated to an individual close to the corresponding author.

\bibliography{cas-refs}
\clearpage

\appendix
\section{Dataset Details}
\label{appendix:dataset}
\subsection{Data Structure}

For each problem, we implemented a structured extraction pipeline converting unstructured textbook content into rigorous evaluation format. 
Each problem is decomposed into two components: 
(1) Problem Description preserving raw context (narrative, operational logic, system schematics), 
and (2) Ground Truth Model with explicit segmentation into decision variables, objective functions, and constraints.

\begin{figure*}
    \centering
    \tcbset{
        enhanced,
        colback=white,
        colframe=black,
        coltitle=white,
        sharp corners,
        boxrule=0.8pt,
        left=4pt, right=4pt, top=4pt, bottom=4pt,
        title style={fill=black}
    }

    \begin{tcbraster}[raster columns=2, raster equal height=rows, raster force size=false, raster column skip=0.02\textwidth]

        \begin{tcolorbox}[title=Problem Description Component]
            \small
            \textbf{\textcolor{darkgreen}{Problem Background \& Objective:}}
            \par\medskip
            Optimize a 4-stage distillation column (reboiler to condenser) to minimize annual operating costs, specifically reboiler heat duty, while meeting strict product purity specs.

            \vspace{0.5em}
            \textbf{\textcolor{darkgreen}{Diagrams:}}
            \par\smallskip
            \centering
            \includegraphics[width=0.9\linewidth, keepaspectratio]{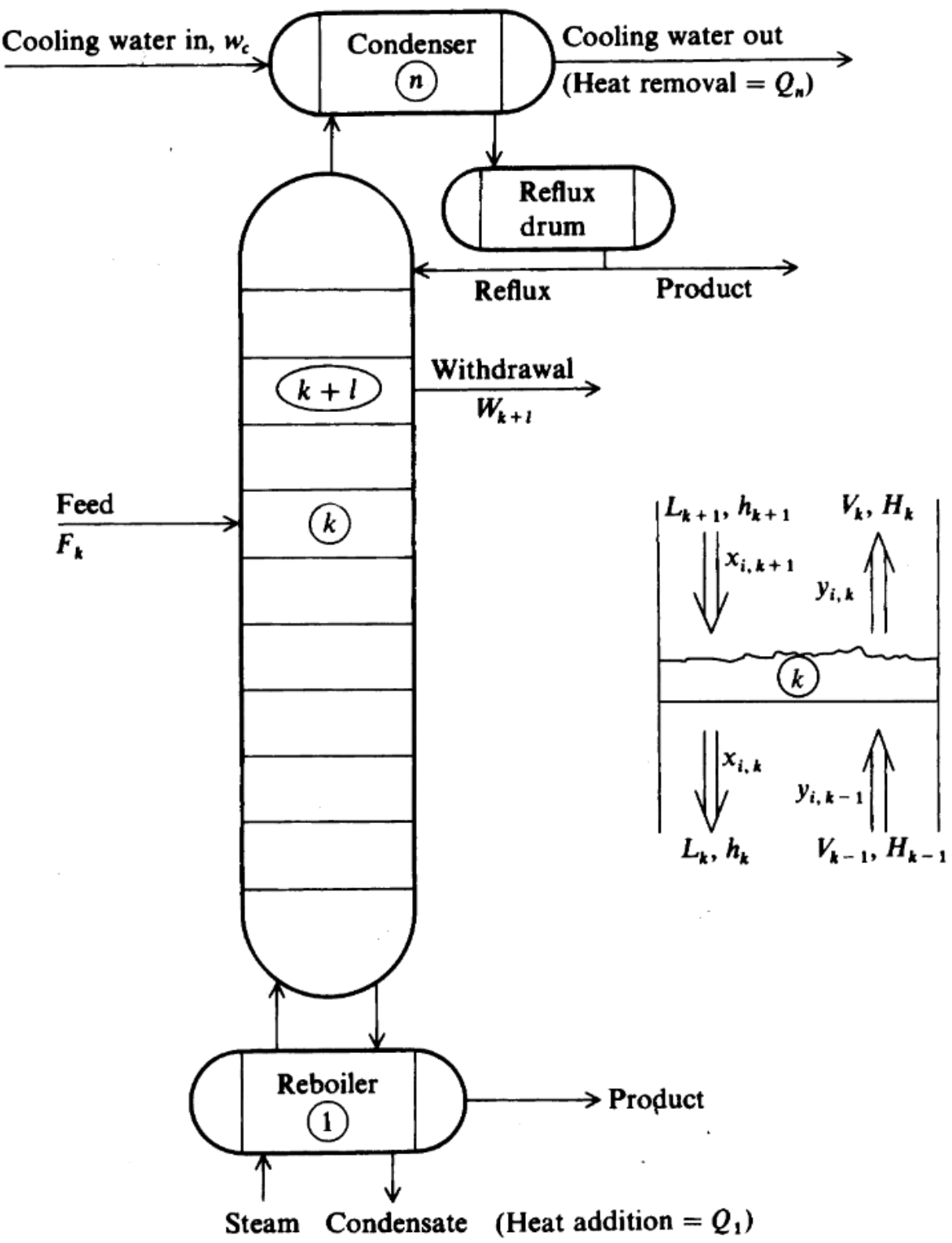}
        \end{tcolorbox}
        \begin{tcolorbox}[title=Ground Truth Model]
\small

            \textbf{\textcolor{darkgreen}{Sets \& Parameters:}} \\
            \textbf{Stages:} $k \in \{1 \dots 4\}$ (1=Reboiler, 4=Condenser) \\
            \textbf{Components:} $i \in \{1 \dots N_c\}$ \\
            \textbf{Specs:} $F_{tot}=100$, $W_{top}=10$ (lb mol/h)

            \rule{\linewidth}{0.4pt}

            \textbf{\textcolor{darkgreen}{Variables (Degrees of Freedom):}} \\
            \textbf{Control:} $u = [Q_1, F_1 \dots F_4]$ \\
            \textbf{State:} $L_k, V_k, x_{i,k}, y_{i,k}, T_k, P_k$

            \rule{\linewidth}{0.4pt}

            \textbf{\textcolor{darkgreen}{Objective Function:}}
            \begin{equation*}
                \min_{u} \quad J = Q_1 \quad
            \end{equation*}

            \text{\small{Minimize Reboiler Heat}}

            \rule{\linewidth}{0.4pt}

            \textbf{\textcolor{darkgreen}{Constraints ($k=1\dots n$):}}
            \small
            \begin{align*}
                 & \textbf{Conservation Laws:} \\
                 & F_k^V + L_{k+1} + V_{k-1} = V_k + L_k + W_k \\[-0.5ex]
                 & \quad \hookrightarrow \text{(Total Mass Bal.)} \\
                 & F_k z_{i,k} + L_{k+1}x_{i,k+1} + \dots = V_k y_{i,k} + \dots \\[-0.5ex]
                 & \quad \hookrightarrow \text{(Comp. Mass Bal.)} \\
                 & Q_k + h_k^F F_k + \dots = H_k V_k + h_k L_k \\[-0.5ex]
                 & \quad \hookrightarrow \text{(Energy Bal.)} \\[0.3em]
                 & \textbf{Physico-Chemical:} \\
                 & y_{i,k} = K_{i,k}(T,P) x_{i,k} \quad \text{(Phase Eq.)} \\
                 & \sum\nolimits_i x_{i,k} = 1, \; \sum\nolimits_i y_{i,k} = 1 \\
                 & H_k = f^V(T,P,y), \; h_k = f^L(T,P,x) \\[0.3em]
                 & \textbf{Bounds \& Specs:} \\
                 & x_{i,\text{prod}}^{min} \leq x_{i,\text{prod}} \leq x_{i,\text{prod}}^{max} \quad \text{(Quality)} \\
                 & L_k, V_k, x_{i,k}, y_{i,k} \geq 0
            \end{align*}
        \end{tcolorbox}

    \end{tcbraster}

    \caption{Example of a structured benchmark sample (Distillation Optimization). The \textbf{left} panel contains the unstructured problem context, while the \textbf{right} panel shows the expert-verified Ground Truth model.}
    \label{fig:data_structure_distillation}
\end{figure*}

To establish a rigorous evaluation environment for \systemname, IPC-Bench was carefully curated from seminal process control and chemical engineering optimization textbooks.
Unlike traditional mathematical reasoning benchmarks that rely on simplified, single-modality textual puzzles evaluated via direct localized execution, our dataset is designed to reflect authentic industrial complexities.
For each problem, we implemented a structured extraction pipeline that translates unstructured, multimodal engineering content into a rigorous, machine-evaluable representation.
Each problem is strictly decomposed into two distinct components: 
(1) Problem Description Component: Preserves the raw industrial context. 
This includes the textual narrative detailing operational logic, system objectives, and boundaries, as well as multimodal elements like Piping and Instrumentation Diagrams (P\&IDs) or system schematics, which are essential for inferring spatial and topological relationships.
(2) Ground Truth Model: Presents the mathematical formulation with explicit semantic segmentation. 
Rather than a flat list of equations, the model is strictly categorized into variables, objective functions, and constraints.
An example is illustrated in \cref{fig:data_structure_distillation}.

\section{Knowledge Base Structure}
\label{appendix:kb}

The current knowledge base contains 55 self-contained retrieval entries. 
Rather than organizing knowledge as broad optimization modules or isolated algorithm cards, it represents reusable problem pathways.
General entries capture structures that recur across process industries, while contextualized entries adapt the same design to concrete industrial settings using anonymized patterns distilled from our industrial project experience. 
Each entry is indexed and retrieved as an independent context unit and links an observable symptom or operating context to a primary optimization objective, conditional candidate methods, required process information, reusable mathematical and constraint patterns, and applicability boundaries.
The symptom serves as a retrieval anchor rather than a confirmed root cause, and the method field records selection conditions rather than a unique recommendation. The knowledge base remains a contextual engineering repository rather than an executable physics engine, and its content does not automatically enforce physical laws or establish solver feasibility.

At runtime, vector search and a cross-encoder reranker select entries based on the structured process context and current objective. 
The objective and method fields guide Strategy Design; information requirements expose missing inputs; and formulation patterns and boundaries support Mathematical Modeling and subsequent human review. Source provenance is maintained separately from the retrieval text so that individual entries remain self-contained while their evidence and maintenance history can be audited.
\Cref{tab:kb_example} illustrates this design with an anonymized evaporator concentration-control example.

\begin{table*}
\centering
\caption{Representative contextualized knowledge entry for an evaporator concentration-control scenario.}
\label{tab:kb_example}
\small
\begin{tabularx}{0.99\linewidth}{>{\raggedright\arraybackslash}p{0.21\linewidth} >{\raggedright\arraybackslash}X}
\toprule
\textbf{Entry field} & \textbf{Representative content} \\
\midrule
\textbf{Operating context and symptom} & Product concentration varies after changes in feed flow or composition; delayed measurements lead to repeated steam adjustments and overshoot. \\
\midrule
\textbf{Optimization objective} & Stabilize product concentration while respecting product-quality and steam-system limits and avoiding unnecessary energy use. \\
\midrule
\textbf{Conditional candidate methods} & Use feedforward plus feedback when feed disturbances are measured reliably; use state estimation and MPC when a validated dynamic model and active operating constraints are available; use a slower economic optimization layer only when the task is to update economically preferred operating targets. \\
\midrule
\textbf{Required process information} & Feed flow and composition, product-concentration measurements and delay, steam flow and pressure, sampling interval, candidate manipulated variables, operating bounds, and historical input--output responses. \\
\midrule
\textbf{Mathematical and constraint patterns} & Material and energy balances, delayed state transitions, input and output bounds, move-rate constraints, measurement equations, and terminal or tracking objectives. \\
\midrule
\textbf{Applicability boundary} & Missing dynamics, actuator limits, quality bounds, and measurement timing remain unresolved rather than being inferred. Site-specific safety and product requirements require engineering confirmation and subsequent numerical validation. \\
\bottomrule
\end{tabularx}
\end{table*}


\begin{figure*}
    \centering
    \includegraphics[width=0.7\linewidth]{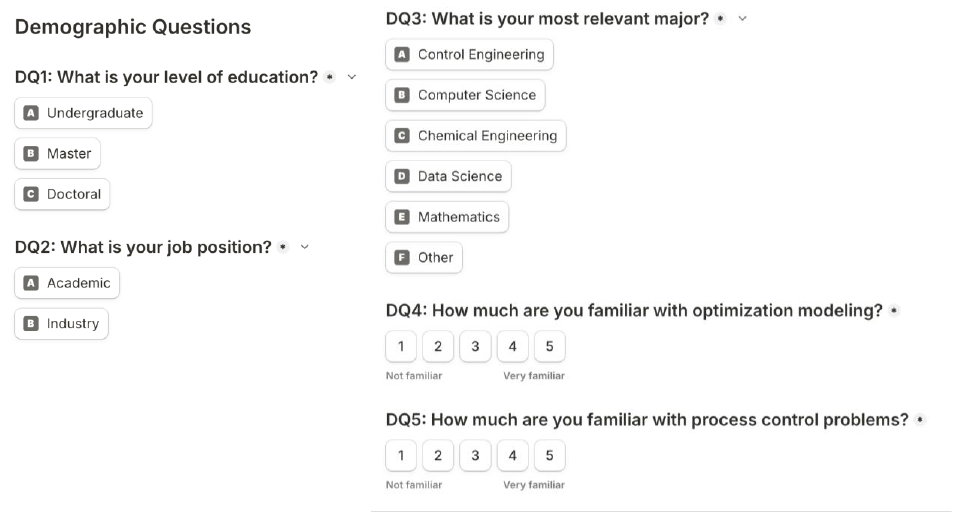}
    \caption{Demographic and expertise profiling questions for human evaluators.}
    \label{fig:demoques}
\end{figure*}

\section{Human Evaluation Survey Design}
\label{appendix:survey}

The human evaluation was conducted via a structured survey instrument designed to capture both evaluator expertise profiles and technical assessments of the generated models.
The evaluation panel consisted of 8 domain specialists (Ph.D. students and postdocs) from computer science, data science, and control engineering.

As illustrated in \cref{fig:demoques}, the instrument begins with a demographic section profiling evaluator expertise: highest educational attainment, current professional role (Academic vs. Industry), and self-reported familiarity with optimization modeling and industrial processes on a 5-point Likert scale. This profiling ensures technical judgments are rendered by qualified individuals.

\begin{figure*}
    \centering
    \includegraphics[width=0.8\linewidth]{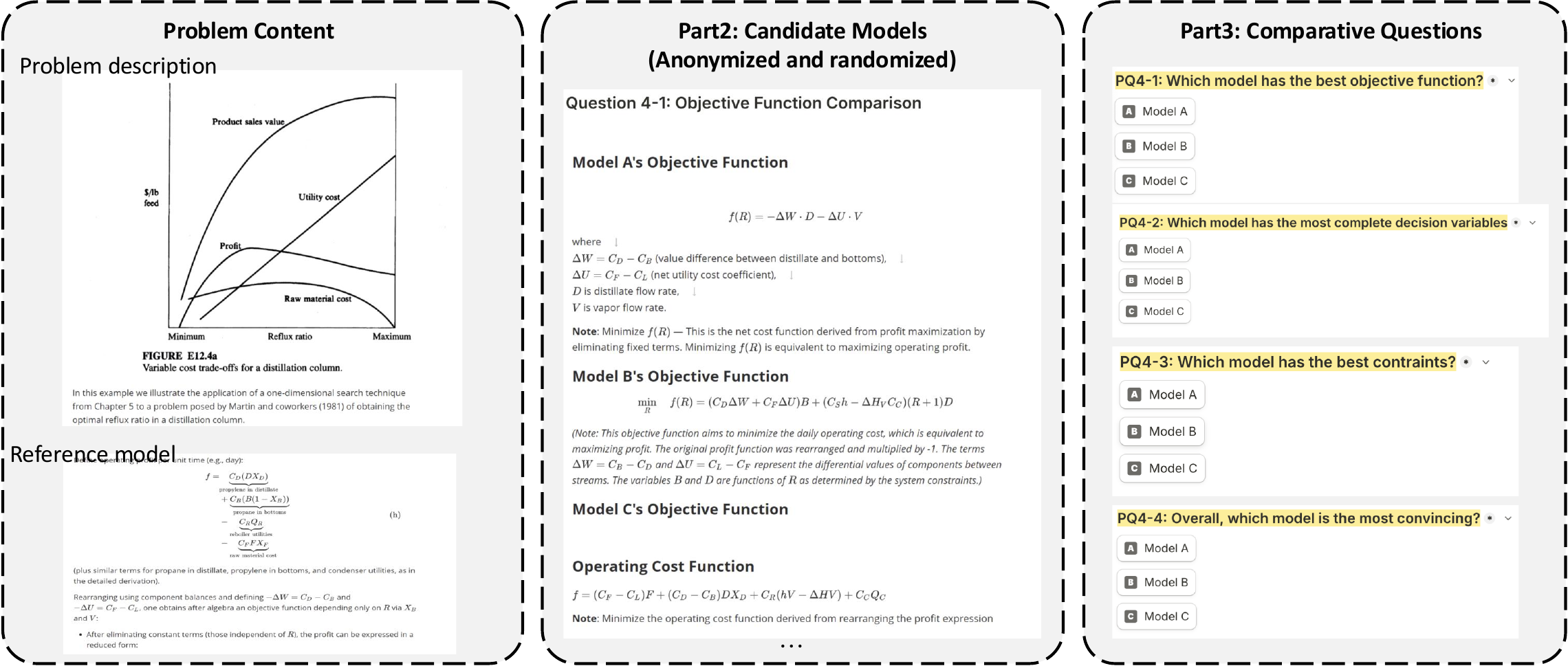}
    \caption{Comparative assessment interface for each test case in the human evaluation.}
    \label{fig:evalflow}
\end{figure*}

Following demographic profiling, experts were presented with randomized test cases as shown in \cref{fig:evalflow}. For each problem, the interface displays the original multimodal problem description alongside a reference ground truth model and a blinded set of candidate models generated by \systemname{} and baseline LLMs. Experts performed a comparative assessment across four dimensions: (1) Objective Function Quality, (2) Decision Variable Completeness, (3) Constraint Validity, and (4) Overall Convincingness.

\paragraph{Evaluation Panel}
To ensure a balanced assessment of both computational and physical aspects, the panel comprised experts from complementary disciplines: 3 from computer science, 2 from data science, and 3 from control engineering.
The panel's qualifications were verified through self-assessment of domain familiarity, summarized in \cref{tab:expert_demographics}.
For Optimization Modeling, the majority rated themselves at high familiarity levels (4--5 on a 5-point Likert scale), ensuring rigorous judgment of mathematical formulations.

\begin{table}
    \centering
    \small
    \setlength{\tabcolsep}{4pt}
    \caption{Distribution of Expert Self-Assessment on Domain Familiarity (Scale 1--5)}
    \label{tab:expert_demographics}
    \begin{tabular}{lccccc}
    \toprule
    \textbf{Domain / Familiarity} & \textbf{1 (Low)} & \textbf{2} & \textbf{3} & \textbf{4} & \textbf{5 (High)} \\ \cmidrule(lr){1-6}
    Optimization Modeling & 0 & 0 & 2 & 4 & 2 \\
    Process Control & 0 & 2 & 4 & 2 & 0 \\
    \bottomrule
    \end{tabular}
\end{table}

\paragraph{Bias Control and Blinding}
To mitigate potential bias, all generated models underwent a strict anonymization process where references to specific agents or underlying model architectures were removed. The presentation order of models (Model A, B, C) was randomized within each question to prevent position bias. The original problem statement was always provided alongside the models as a unified reference for verification.

\paragraph{Statistical Analysis}
We employed a two-sided exact binomial test with null hypothesis $H_0$ that expert preference follows a random distribution ($p = 1/3$), and alternative hypothesis $H_1$ that experts prefer \systemname{} more frequently ($p > 1/3$). Significance was determined at the $\alpha = 0.05$ level. Inter-rater reliability was assessed qualitatively by analyzing preference variance across expertise levels to ensure robustness across the evaluator population.

\section{Machine Evaluation Protocol}
\label{appendix:machine-eval}

To complement human-expert assessment with scalable, reproducible metrics, we developed an automated evaluation pipeline that quantitatively compares the predicted model $M_p = \{V_p, f_p, C_p\}$ against the ground truth model $M_g = \{V_g, f_g, C_g\}$ across three dimensions.

\paragraph{Decision Variables}
Variables are treated as discrete sets. A regex module normalizes variables by stripping spaces, standardizing subscripts, and enforcing lowercase (e.g., $x_1 \to x1$). An LLM-driven semantic mapping function then aligns differently named but physically equivalent entities using natural language descriptions. The structural fidelity is calculated via Jaccard similarity:
\begin{equation}
    S_{vars} = \frac{|V_g \cap V_p'|}{|V_g \cup V_p'|}
\end{equation}
A high $S_{vars}$ indicates close set-level agreement with the reference variables, subject to the quality of the semantic mapping.

\paragraph{Objective Function}
The evaluation script uses regular expressions to strip natural language noise (prefixes like ``Minimize'', ``Max'', ``f(x)''). The score is the Jaccard similarity between token sets:
\begin{equation}
    S_{obj} = \frac{|T_g \cap T_p|}{|T_g \cup T_p|}
\end{equation}

\paragraph{Constraint Set}
All terms are moved to the left side (e.g., $A \le B \to A - B \le 0$). A weight matrix $w_{ij}$ captures mathematical token similarity between each generated constraint $c_{p,j}$ and ground truth constraint $c_{g,i}$. Maximum bipartite matching (Hungarian algorithm) pairs each generated constraint with its closest ground truth equivalent. Scoring uses a continuous F1 metric:
\begin{equation}
    P = \frac{\sum w_{ij}}{|C_p|}, \quad R = \frac{\sum w_{ij}}{|C_g|}, \quad S_{cons} = 2 \cdot \frac{P \cdot R}{P + R}
\end{equation}
Precision $P$ penalizes hallucinated constraints; recall $R$ penalizes missing safety interlocks or conservation laws.

\paragraph{Human-Machine Alignment}
We validated the automated metrics against expert judgment by treating the human-preferred model as ground truth. For each of the 20 evaluated cases, the machine pipeline ranked candidate models. Alignment was quantified using Top-1 Accuracy (frequency the machine's highest-scoring model matches the human choice) and Mean Reciprocal Rank: $\text{MRR} = \frac{1}{N} \sum_{i=1}^{N} \frac{1}{\text{rank}_i}$, where $\text{rank}_i$ is the position of the human-preferred model in the machine-sorted list for case $i$.

\newpage
\section{Case Study}
\label{appendix:cases}

To make the intermediate representations in \cref{fig:workflow} concrete, we present an abridged trace for a mixing-tank optimal tracking-control case from the held-out IPC-Bench test set.
The case contains only continuous variables, includes a process diagram as part of its input, and directly matches the process-control scope of this work.
The trace preserves the recorded information-structuring step, the subsequent clarification of the dynamic task, and the final formulation while shortening conversational acknowledgements.
The diagram supplies the nominal inlet and outlet conditions and the two control setpoints; the interaction then confirms the dynamic energy balance, tracking objective, and inlet-flow bounds.
Parameters that remain unspecified---including the tank volume, objective weights, flow limits, time horizon, and initial temperature---are retained symbolically rather than assigned invented values.
The resulting artifact is a complete parameterized optimal-control model organized according to the same decision-variable, objective-function, and constraint categories used in evaluation.

\begin{figure*}[p]
    \centering
    \tcbset{
        enhanced,
        sharp corners,
        boxrule=0.7pt,
        left=5pt, right=5pt, top=4pt, bottom=4pt,
        colframe=black,
        colback=white,
        colbacktitle=white,
        coltitle=black,
        fonttitle=\bfseries\small,
        titlerule=0.5pt
    }

    \begin{tcolorbox}[title={1. Case input $\mathcal{X}$}]
        \scriptsize
        \begin{minipage}[c]{0.54\linewidth}
            A perfectly mixed tank receives a hot stream and a cold stream.
            Choose their flow rates so that the total outlet flow and tank temperature
            track their respective setpoints. Density and heat capacity are constant,
            and the hotter inlet satisfies $T_1^{\circ}>T_2^{\circ}$.

            \medskip
            The diagram gives nominal conditions
            $F_1=20\,\mathrm{kg/min}$, $T_1^{\circ}=434\,\mathrm{K}$,
            $F_2=40\,\mathrm{kg/min}$, and $T_2^{\circ}=293\,\mathrm{K}$,
            together with $F_{\mathrm{set}}=60\,\mathrm{kg/min}$ and
            $T_{\mathrm{set}}=340\,\mathrm{K}$.
        \end{minipage}\hfill
        \begin{minipage}[c]{0.42\linewidth}
            \centering
            \includegraphics[width=\linewidth,keepaspectratio,trim=0 45 0 0,clip]{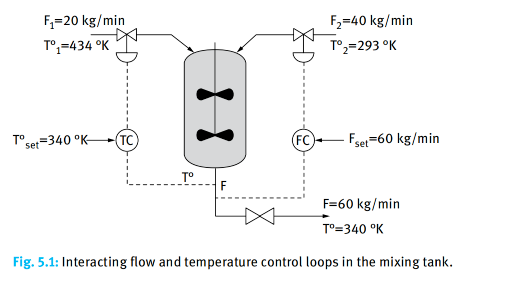}
        \end{minipage}
    \end{tcolorbox}

    \vspace{-0.45em}
    {\Large$\Downarrow$}
    \vspace{-0.45em}

    \begin{tcolorbox}[title={2. Information Structuring output $S_0$}]
        \scriptsize
        \begin{tabularx}{\linewidth}{@{}p{0.19\linewidth}X@{}}
            \textbf{System boundary} & One well-mixed tank, two inlet streams, and one outlet; the tank temperature equals the outlet temperature. \\
            \textbf{Candidate decisions} & Inlet flow rates $F_1$ and $F_2$; inlet temperatures $T_1^{\circ}$ and $T_2^{\circ}$ are fixed process data. \\
            \textbf{Control goal} & Track both the total outlet-flow setpoint $F_{\mathrm{set}}$ and temperature setpoint $T_{\mathrm{set}}$. \\
            \textbf{Clarification needed} & Determine whether the task is steady-state or dynamic, and obtain the applicable balance equations, horizon, weights, bounds, and initial condition. \\
        \end{tabularx}
    \end{tcolorbox}

    \vspace{-0.45em}
    {\Large$\Downarrow$}
    \vspace{-0.45em}

    \begin{tcolorbox}[title={3. Strategy Design output $Z$}]
        \scriptsize
        The clarification confirms a dynamic optimal-tracking formulation.
        Optimize the continuous control profiles $F_1(t)$ and $F_2(t)$ over
        $t\in[0,t_f]$; use the tank energy balance as the state equation;
        penalize accumulated outlet-flow and temperature tracking errors; and
        enforce bounds on both inlet flows. Treat $V$, $w_1$, $w_2$, $t_f$,
        the flow limits, and $T_0$ as parameters to be supplied rather than inferred.
    \end{tcolorbox}

    \vspace{-0.45em}
    {\Large$\Downarrow$}
    \vspace{-0.45em}

    \begin{tcolorbox}[title={4. Mathematical Modeling output $\mathcal{M}$}]
        \scriptsize
        \setlength{\abovedisplayskip}{3pt}
        \setlength{\belowdisplayskip}{3pt}

        \textbf{Decision Variables.}
        \begin{equation*}
            F_1(t),\;F_2(t), \qquad t\in[0,t_f].
        \end{equation*}
        $F_1(t)$ and $F_2(t)$ are the hot- and cold-inlet flow-rate profiles selected by the optimizer.
        The state $T(t)$ is the well-mixed tank temperature, and
        $F(t)$ is the derived total outlet flow; neither is an independent decision variable.

        \textbf{Objective Function.}
        \begin{equation*}
            \min_{F_1(\cdot),F_2(\cdot)}\quad
            J=\int_{0}^{t_f}\!\left[
            w_1\bigl(F(t)-F_{\mathrm{set}}\bigr)^2
            +w_2\bigl(T(t)-T_{\mathrm{set}}\bigr)^2
            \right]dt.
        \end{equation*}
        $F_{\mathrm{set}}=60\,\mathrm{kg/min}$ and $T_{\mathrm{set}}=340\,\mathrm{K}$
        are the desired outlet conditions; nonnegative weights $w_1$ and $w_2$
        set the relative importance of flow and temperature tracking.

        \textbf{Constraints.}

        \emph{Temperature dynamics:}
        \begin{equation*}
            \frac{dT(t)}{dt}
            =\frac{F_1(t)T_1^{\circ}+F_2(t)T_2^{\circ}
            -\bigl(F_1(t)+F_2(t)\bigr)T(t)}{V}.
        \end{equation*}
        This energy balance determines the temperature trajectory; $V$ is the
        constant tank volume and $T_1^{\circ},T_2^{\circ}$ are fixed inlet temperatures.

        \emph{Initial condition and outlet-flow definition:}
        \begin{align*}
            T(0)&=T_0,\\
            F(t)&=F_1(t)+F_2(t).
        \end{align*}
        $T_0$ initializes the dynamic state, while the second equality is the
        constant-density total mass balance.

        \emph{Control bounds:}
        \begin{equation*}
            F_i^{\min}\leq F_i(t)\leq F_i^{\max},
            \qquad i\in\{1,2\},\quad t\in[0,t_f].
        \end{equation*}
        $F_i^{\min}$ and $F_i^{\max}$ denote the admissible operating range of each inlet valve.
    \end{tcolorbox}

    \caption{Abridged end-to-end trace for a mixing-tank optimal tracking-control case from the held-out IPC-Bench test set. The input includes the process diagram; Information Structuring identifies the system boundary, candidate controls, and missing specifications; Strategy Design incorporates the clarified dynamic task; and Mathematical Modeling produces the decision variables, objective, and constraints used by the evaluation protocol. Unspecified numerical parameters remain explicit.}
    \label{fig:mixing_tank_trace}
\end{figure*}

\clearpage

\section{Prompt Engineering Details}
\label{appendix:prompts}

We implemented a modular prompting strategy with role-constrained instructions for each stage.
The three stages progress from organizing available facts, to proposing a high-level control strategy, and finally drafting a candidate mathematical specification for review.

\subsection{Stage 1: Information Assessment}


This stage functions as an intake filter that transforms unstructured, multimodal engineering descriptions into a coherent, structured representation before any reasoning begins. 
Our core design insight here is to enforce a rigorous ``gap-analysis'' protocol. 
By classifying gathered information into confidence tiers (e.g., confirmed data, partially known information requiring clarification, and critical missing data), this triage mechanism makes unsupported assumptions more visible to downstream stages and reviewers.

\begin{promptbox}[label={prompt:info_struct}, title={Stage 1: Information Assessment Prompt}]

    [Role]

    You are a professional Industrial Optimization Expert specializing in solving complex industrial modeling and optimization problems.

    [Objective]

    Receive raw materials about an industrial optimization problem and transform them into a comprehensive, structured ``Complete Information Summary Report'' through a single-pass analysis.

    [Execution Rules]

    - Comprehensive Analysis: Extract and interpret all information provided in the user's raw material, including textual narratives and visual process diagrams.

    - Identification and Inference: Identify the optimization type, core objectives, adjustable variables, system constraints, KPIs, and target values. Make reasonable inferences based on the provided material where information is implicit.

    - Gap Analysis: Mark the completeness of every extracted item using:

    \quad {\textbf{(\ding{51})}} Information sufficient for modeling

    \quad {\textbf{($\triangleright$)}} Information partially missing (requires clarification)

    \quad \textbf{(\ding{55})} Information critically missing

    - Structured Output: Generate a report with the following seven sections:

    \quad 1) Optimization Type and Core Objective

    \quad 2) Adjustable Variables (with current values, ranges, difficulty, and impact)

    \quad 3) System Constraints (physical, safety, regulatory, operational)

    \quad 4) KPI Priority and Target Values

    \quad 5) Domain-Specific Information (process steps, equipment, data points, workflows)

    \quad 6) Historical Optimization Experience

    \quad 7) Information Gaps and Their Potential Impact on Modeling

    [Mandatory Interaction]

    ``Does this information accurately reflect your situation? Do you need to add anything?''
\end{promptbox}

\newpage
\subsection{Stage 2: Strategy Design}


This stage acts as the reasoning core that bridges the initial information assessment and the eventual mathematical formulation. 
The prompt design emphasizes a multi-dimensional analysis approach—exploring aspects like process flow bottlenecks, equipment efficiency, and energy utilization—so the agent can autonomously select the single most promising optimization strategy rather than offering a generic menu of options. 
A key feature of this design is the three-level drill-down: forcing the model to explicitly link core problems to their production-floor manifestations, and ultimately to their root physical or operational causes. 
A mandatory confirmation gate lets the user correct the proposed high-level strategy using tacit operational knowledge before mathematical syntax is generated.

\begin{promptbox}[label={prompt:strategy_design}, title={Stage 2: Strategy Design Prompt}]

    [Role]

    You are a professional Industrial Optimization Expert. Your task is to execute the ``System Problem Analysis and Optimization Strategy Determination'' sub-task.

    [Input]

    - Complete Information Summary Report (from Stage~1)

    - Retrieved Knowledge Base entries (as context)

    [Objective]

    Conduct an in-depth multi-dimensional analysis of the system and autonomously determine the single optimal optimization strategy to prepare for mathematical modeling.

    [Execution Rules]

    - Data Sourcing: Explicitly list the key data points from the summary report that this analysis relies on.

    - Multi-Dimensional Analysis: Analyze the system from at least three distinct dimensions (e.g., process flow bottlenecks, energy utilization efficiency, equipment constraints). For each dimension, perform a three-level breakdown:

    \quad 1) \textbf{Core Problem}: A precise problem statement grounded in the reported data.

    \quad 2) \textbf{Primary Manifestations}: Specific ways the problem appears in production, described with observable data or phenomena.

    \quad 3) \textbf{Root Cause Analysis}: The fundamental physical or operational causes.

    - Strategy Decision: Identify multiple potential optimization directions, evaluate them on expected impact, implementation feasibility, and data availability, then directly select the single best strategy. Do not present options to the user.

    - Scope Restriction: Strictly limit analysis to System Analysis, Mathematical Modeling, and Algorithm Design. Reject out-of-scope requests.

    [Output Format]

    A structured report containing: (1) Key data points used, (2) Multi-dimensional problem analysis with the three-level drill-down, (3) Evaluation of potential directions, (4) The final selected strategy with justification (impact, feasibility, alignment with objectives).

    [Mandatory Interaction]

    ``Do you agree with this optimization strategy? We need your confirmation to proceed to mathematical modeling.''
\end{promptbox}

\newpage
\balance
\subsection{Stage 3: Mathematical Formulation}


The final reasoning stage translates the acknowledged optimization strategy into a candidate mathematical formulation.
The foundational design principle is data-variable traceability: the instructions request that every defined symbol, whether a decision variable or a fixed parameter, be linked to a data point or marked as unresolved.
This makes ungrounded variables easier to detect but does not establish physical or numerical validity.
The prompt also requests analysis of linearity, convexity, scale, data dependency, and likely solver family so that a human can assess implementation requirements.

\begin{promptbox}[label={prompt:modeling}, title={Stage 3: Mathematical Modeling Prompt}]

    [Role]

    You are a professional Industrial Optimization Expert. Your task is to execute the mathematical modeling sub-task based on the confirmed optimization strategy.

    [Input]

    - Complete Information Summary Report (from Stage~1)

    - System Analysis and Optimization Strategy Report (from Stage~2)

    [Objective]

    Translate the confirmed strategy into a candidate mathematical optimization specification for engineering review and subsequent solver implementation.

    [Execution Rules]

    - Data Sourcing: Detail all key data points from the summary report used in this model.

    - Data-Variable Mapping: Every defined variable and parameter must link to a specific data point from the Stage~1 summary report. Present in table format:

    \quad \textbf{Symbol | Meaning | Unit | Data Source}

    - Model Justification: Explicitly state why the chosen model class (e.g., MILP, MPC, NLP) was selected based on system characteristics and the confirmed strategy.

    - Mathematical Expression: Use LaTeX syntax for the objective function and all constraint equations.

    - Constraint Categorization: Classify all constraints into:

    \quad \textbf{Physical Constraints}: Mass/energy balances, thermodynamic limits, phase equilibria.

    \quad \textbf{Operational Constraints}: Equipment capacity bounds, safety interlocks, quality specifications.

    \quad \textbf{Logical Constraints}: Binary on/off states, sequencing requirements, conditional rules.

    - Model Characteristics Analysis: Assess (1) solving difficulty (linearity, convexity, scale), (2) data dependency (which inputs drive model sensitivity), and (3) expected outcomes (quantified improvements).

    [Output Format]

    A structured ``Mathematical Model Specification'' containing: variable/parameter definition table, objective function in LaTeX, constraints organized by category, and model characteristics analysis.

\end{promptbox}


\end{document}